\documentclass[11pt]{article}

\usepackage[margin=1in]{geometry}
\usepackage{amsmath,amssymb}
\usepackage{booktabs}
\usepackage{array}
\usepackage[hypertexnames=false]{hyperref}
\usepackage[numbers,sort&compress]{natbib}
\usepackage{xcolor}
\usepackage{listings}
\usepackage{float}
\usepackage{graphicx}
\usepackage{tikz}
\usetikzlibrary{positioning,arrows.meta,fit}

\hypersetup{
  hidelinks,
  pdftitle={Clinical Knowledge Graphs for Chest X-Ray Device Reasoning},
  pdfauthor={Harshil Lodhiya},
  pdfkeywords={chest x-ray, clinical knowledge graph, catheter placement, uncertainty, radiology reports, multimodal AI}
}
\lstdefinestyle{pseudo}{basicstyle=\ttfamily\small,breaklines=true,frame=single,columns=fullflexible,keepspaces=true}

\title{Clinical Knowledge Graphs\\
for Chest X-Ray Device Reasoning}

\author{Harshil Lodhiya\\Sliced Health\\\texttt{hlodhiya@slicedhealth.com}}
\date{August 2026}

\begin{document}
\maketitle

\begin{abstract}
Chest radiographs are routinely used to check the position of catheters, tubes, and other support devices. The relevant evidence, however, is scattered across images, reports, placement labels, procedures, and subsequent studies. Image models usually return a label or segmentation, while report systems structure text. Neither alone can support an auditable question such as whether a particular device remained malpositioned, whether a recommendation was followed, or whether the available evidence is too uncertain for an automated conclusion.

We describe a clinical knowledge graph for chest X-ray device reasoning. Its implemented visual path turns an image into device instances, tip-location distributions, placement scores, and fragment-level provenance, then writes those observations into a typed evidence graph. A report-event path is specified for use with report-bearing data; it retains source spans, assertion status, recommendations, and temporal cues. Uncertainty is kept with the evidence: tip covariance captures geometric uncertainty, while extraction, grounding, and cross-study links carry their own uncertainty attributes.

Using saved predictions from the complete RANZCR CLiP test archive (30,083 studies from 3,255 patients across five non-overlapping outer folds), the graph builder materialized 914,632 B7 evidence nodes and 884,549 typed relationships. Every one of the 118,647 B7 predicted-device nodes included a tip covariance, placement-probability distribution, fragment provenance, and fragment count; the direct B2 baseline included none of these fields. This is a post-hoc descriptive analysis of already-saved test predictions, since the original test-plan lock was not enabled before they were generated. It shows that the visual evidence graph can be built reproducibly at scale. It does not establish report grounding, longitudinal reasoning, clinical benefit, or prospective utility. We therefore describe the multimodal extensions and the evaluation needed to test them.
\end{abstract}


\nocite{johnson2019mimiccxr}
\nocite{irvin2019chexpert}
\nocite{ranzcr2021clip}
\nocite{lodhiya2026ucompcxr}
\nocite{jain2021radgraph}
\nocite{kendall2017uncertainties}
\nocite{gal2016dropout}
\nocite{lakshminarayanan2017ensembles}
\nocite{lodhiya2026eventforge}
\nocite{guo2017calibration}
\nocite{lewis2020rag}
\nocite{edge2024graphrag}
\nocite{lodhiya2026schemaaware}
\nocite{sarthi2024raptor}
\nocite{khattab2023dspy}
\nocite{hagberg2008networkx}
\nocite{wang2017chestxray8}

\section{Introduction}

Catheters and tubes are common in hospitalized patients. Endotracheal tubes secure airways, nasogastric tubes support feeding or decompression, central venous catheters enable medication delivery and monitoring, and Swan-Ganz catheters support hemodynamic assessment. After placement or repositioning, chest radiography is often used to verify location. The clinical question is not merely whether a device is present. The relevant question is whether a particular device instance is correctly positioned, whether the tip is confidently localized, whether the report recommends action, and whether a later study or procedure resolved the issue.

Current computational systems capture only part of this clinical structure. Image-level classifiers can predict labels such as ``ETT abnormal'' or ``CVC borderline'', but they generally do not identify which device instance produced the label. Segmentation systems can trace device pixels, but they may collapse multiple overlapping devices into one mask and may not express calibrated uncertainty at the tip. Report extraction systems can identify radiology findings, but text alone cannot supply device geometry or spatial uncertainty. Longitudinal clinical reasoning requires all of these signals together.

We use an uncertainty-aware clinical knowledge graph to connect device instances, tip estimates, placement labels, report findings, recommendations, studies, encounters, procedures, and temporal relations. This makes it possible to ask questions that neither image AI nor report retrieval can answer well on its own:

\begin{itemize}
  \item Did the same CVC remain borderline or abnormal across consecutive studies?
  \item Which report recommendation corresponds to a later repositioning event?
  \item Was a nasogastric tube classified as abnormal because the tip was outside the expected abdominal region or because the image was incomplete?
  \item Which cases had high geometric uncertainty and should be prioritized for radiologist review?
  \item How often did report language conflict with image-derived placement estimates?
\end{itemize}

The design draws on two earlier components. UCompCXR detects local catheter fragments, groups them into device instances, fuses fragment-level tip predictions with precision-weighted Gaussian estimation, and assigns placement labels per device. On RANZCR CLiP, it reported more detected devices and fewer false positives than a multi-task baseline with the same MobileNetV3 backbone, with 95\% tip coverage of 0.948 \citep{lodhiya2026ucompcxr}. EventForge is a PDF-to-knowledge-graph pipeline that extracts typed events, removes near duplicates with embedding similarity, and stores graph and retrieval artifacts in PostgreSQL/pgvector \citep{lodhiya2026eventforge}. Here, the visual uncertainty from the former is paired with the event-graph storage pattern of the latter.

\paragraph{Contributions.}
The paper has five contributions:

\begin{enumerate}
  \item It frames catheter and tube assessment as a multimodal graph problem rather than only an image-classification task.
  \item It defines a schema that links image-derived device instances and tip geometry with report spans, recommendations, procedures, and encounters.
  \item It carries fragment-level uncertainty through to device-level graph evidence and temporal links.
  \item It outlines a report-image grounding strategy that can leave uncertain matches unresolved.
  \item It reports a reproducible visual evidence-graph materialization study and sets out the evaluation required for the remaining multimodal components.
\end{enumerate}

\section{Clinical Motivation}

Chest X-ray device assessment is safety-critical because small placement errors can have meaningful consequences. An endotracheal tube advanced too far can enter a mainstem bronchus; a central venous catheter tip can sit too deep or in an unintended vessel; a nasogastric tube may coil or fail to pass below the diaphragm. The interpretation is also longitudinal. A radiologist may recommend retraction or advancement; a later film may confirm correction; a procedure note may document repositioning. A knowledge graph can represent this trajectory explicitly.

The RANZCR CLiP challenge focused attention on catheter and tube placement labels at scale, including device traces for a subset of images \citep{ranzcr2021clip}. Broader chest X-ray datasets such as ChestX-ray14, CheXpert, and MIMIC-CXR established large-scale image and report resources \citep{wang2017chestxray8,irvin2019chexpert,johnson2019mimiccxr}. However, common label extraction and classification workflows still flatten rich clinical events into image-level labels. This flattening is useful for benchmark training, but it loses the structure needed for patient-level reasoning.

Radiology report graphs partly address the text side. RadGraph, for example, represents entities and relations from radiology reports, enabling structured report understanding \citep{jain2021radgraph}. Yet report graphs alone do not know the image geometry. They can encode that a tube is malpositioned, but not the covariance ellipse around its tip, the competing device instance it may be confused with, or the image-derived uncertainty that should trigger review. The proposed clinical knowledge graph is designed to join these modalities.

\section{System Overview}

The system has four parts: visual extraction, text extraction, graph construction, and reasoning. Table~\ref{tab:planes} gives the division of work.

\begin{table}[t]
\centering
\caption{Core planes in the uncertainty-aware clinical knowledge graph system.}
\label{tab:planes}
\begin{tabular}{p{0.18\linewidth}p{0.35\linewidth}p{0.36\linewidth}}
\toprule
Plane & Responsibility & Representative outputs \\
\midrule
Visual extraction & Detect and assess device instances in chest X-rays & Device fragments, instance paths, tip mean, tip covariance, placement label \\
Text extraction & Extract report findings, recommendations, and events & Report spans, device mentions, action recommendations, temporal cues \\
Graph construction & Link visual, textual, and longitudinal entities & Device nodes, study nodes, report nodes, procedure nodes, temporal edges \\
Reasoning & Answer clinical and quality-improvement queries & Persistence, correction, uncertainty triage, report-image conflict \\
\bottomrule
\end{tabular}
\end{table}

\begin{samepage}
For each imaging study $s$, the system receives an image $x_s$, optional report text $r_s$, patient and encounter metadata, and any available procedure or follow-up events. The output is a graph update:
\[
\Delta G_s = (V_s, E_s, U_s, P_s),
\]
where $V_s$ are nodes, $E_s$ are edges, $U_s$ are uncertainty attributes, and $P_s$ contains provenance.
\end{samepage}

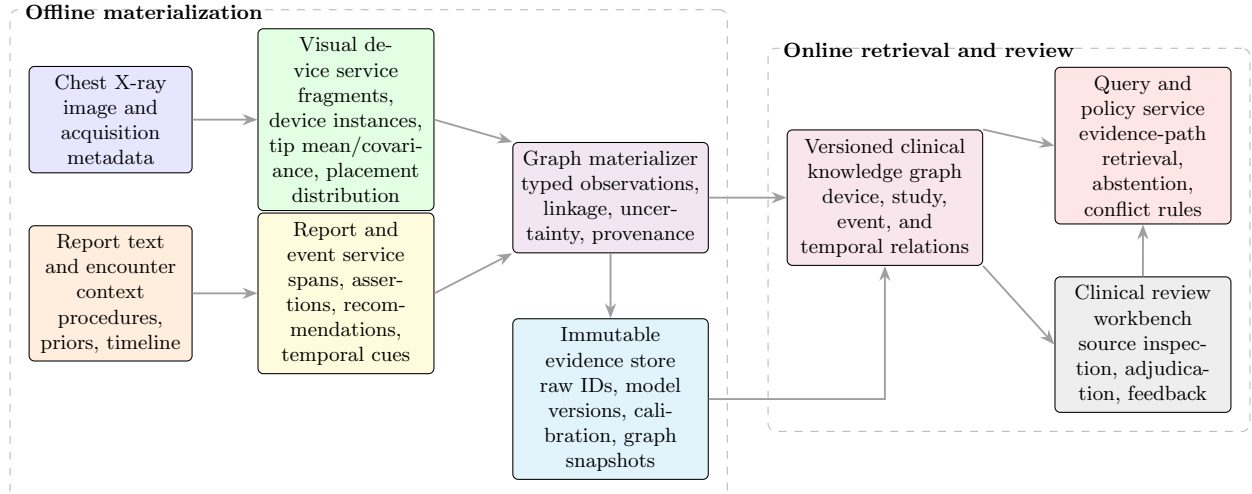
\begin{figure}[H]
\centering
\begin{tikzpicture}[
  scale=0.86,
  transform shape,
  font=\footnotesize,
  source/.style={draw, rounded corners=2pt, align=center, minimum height=1.08cm, text width=2.25cm, fill=#1},
  service/.style={draw, rounded corners=2pt, align=center, minimum height=1.18cm, text width=2.45cm, fill=#1},
  wide/.style={draw, rounded corners=2pt, align=center, minimum height=1.18cm, text width=2.75cm, fill=#1},
  arrow/.style={-{Stealth[length=2.0mm]}, line width=0.7pt, draw=gray!75},
  boundary/.style={draw=gray!55, dashed, rounded corners=3pt, inner sep=7pt}
]
\node[source=blue!10] (image) {Chest X-ray\\image and acquisition metadata};
\node[source=orange!13, below=8mm of image] (textcontext) {Report text and encounter context\\procedures, priors, timeline};

\node[service=green!11, right=10mm of image] (visual) {Visual device service\\fragments, device instances, tip mean/covariance, placement distribution};
\node[service=yellow!16, right=10mm of textcontext] (text) {Report and event service\\spans, assertions, recommendations, temporal cues};

\node[wide=violet!10, right=12mm of visual, yshift=-12mm] (builder) {Graph materializer\\typed observations, linkage, uncertainty, provenance};
\node[wide=cyan!10, below=10mm of builder] (store) {Immutable evidence store\\raw IDs, model versions, calibration, graph snapshots};
\node[wide=purple!10, right=12mm of builder] (graph) {Versioned clinical knowledge graph\\device, study, event, and temporal relations};

\node[service=red!10, right=11mm of graph, yshift=8mm] (api) {Query and policy service\\evidence-path retrieval, abstention, conflict rules};
\node[service=gray!13, below=8mm of api] (review) {Clinical review workbench\\source inspection, adjudication, feedback};

\draw[arrow] (image) -- (visual);
\draw[arrow] (textcontext) -- (text);
\draw[arrow] (visual.east) -- (builder.north west);
\draw[arrow] (text.east) -- (builder.south west);
\draw[arrow] (builder) -- (graph);
\draw[arrow] (builder) -- (store);
\draw[arrow] (store.east) -| (graph.south);
\draw[arrow] (graph.north east) -- (api.west);
\draw[arrow] (graph.south east) -- (review.west);
\draw[arrow] (review.north) -- (api.south);

\node[boundary, fit=(image)(textcontext)(visual)(text)(builder)(store)] (offline) {};
\node[boundary, fit=(graph)(api)(review)] (online) {};
\node[anchor=west, font=\bfseries\footnotesize] at ([xshift=3pt,yshift=-1pt]offline.north west) {Offline materialization};
\node[anchor=west, font=\bfseries\footnotesize] at ([xshift=3pt,yshift=-1pt]online.north west) {Online retrieval and review};
\end{tikzpicture}
\caption{Architecture and deployment flow. Offline services preserve visual, text, and context observations before materializing a versioned evidence graph and immutable provenance store. The online path retrieves evidence from a frozen graph snapshot, applies abstention and conflict policies, and presents source-linked results for human review. The visual service is evaluated in this paper; report and longitudinal services require a report-bearing cohort.}
\label{fig:architecture}
\end{figure}

Figure~\ref{fig:architecture} separates extraction from reasoning. Visual and text components emit attributed observations; the graph builder creates versioned links to those observations. The query layer reads graph snapshots and cannot replace source evidence with an unsupported narrative.

\subsection{Design Requirements}

Five practical rules guide the design. The graph should preserve the identity of a \emph{device instance} instead of collapsing a device family into one image-level label. Each automated assertion needs retrievable evidence: image outputs on the visual side and character offsets plus document version on the text side. Uncertainty should remain uncertainty after graph construction; storing a score must not turn it into a binary fact. Cross-study links should be probabilistic because devices can be replaced, removed, or obscured. Finally, the interface needs an abstention path that sends an uncertain case for review.

This is an evidence model, not a generic document graph. Each node or edge records its producer, model version, input identifier, timestamp, confidence, and validation state. Those fields make later audit possible when a model changes or a report is amended.

\subsection{Data Contracts and Provenance}

Each graph update is checked against a typed contract before it is written. A visual observation includes an image identifier, device-family distribution, spatial coordinate system, placement-score distribution, and model version. A text observation includes the report identifier, a contiguous source span, an assertion category, and a confidence. A derived edge retains the identifiers of the observations used to create it. The point is simple: a downstream conclusion should never outlive the evidence needed to inspect it.

For a graph assertion $a$, provenance is represented as a directed acyclic evidence subgraph $\Pi(a)$. For example, a ``persistent abnormal placement'' assertion points to two placement assessments, their device instances, the underlying images, and a probabilistic cross-study linkage. A user interface can then expose the shortest evidence path rather than a free-text explanation generated after the fact.

\begin{table}[t]
\centering
\caption{Minimum evidence contract for high-value graph assertions. ``Required'' denotes attributes that must be present before a system may expose the assertion as an automated finding.}
\label{tab:contract}
\small
\begin{tabular}{p{0.24\linewidth}p{0.27\linewidth}p{0.34\linewidth}}
\toprule
Assertion & Required evidence & Mandatory provenance attributes \\
\midrule
Device instance & Fragment set or instance mask; family distribution & image UID, model version, coordinate frame, confidence \\
Tip estimate & Mean location and covariance & contributing fragments, fusion rule, calibration set/version \\
Report event & Exact source span; assertion type & report UID, report version, extractor version, confidence \\
Grounded report finding & Text event and candidate device instance & alignment features, threshold, abstention state \\
Persistent malposition & Two assessments and cross-study link & study times, linkage probability, query version \\
Follow-up completed & Recommendation, action, and follow-up evidence & source spans/records, temporal window, matching rule \\
\bottomrule
\end{tabular}
\end{table}

\subsection{Separation of Offline and Online Paths}

The offline path ingests studies, runs extraction, performs candidate linkage, computes calibration summaries, and writes quality-control tables. The online path is narrower: it retrieves precomputed evidence, applies versioned queries, and displays source-linked results. Keeping the two paths separate avoids rerunning high-risk inference at query time and makes retrospective analyses easier to reproduce. It also limits reprocessing to the artifacts affected by a changed model, ontology, or report version.

\section{Visual Device Module}

The visual module follows the UCompCXR design \citep{lodhiya2026ucompcxr}. Given a chest X-ray $x$, it predicts a set of device instances:
\[
D = \{d_1,\ldots,d_K\},
\]
where each instance has
\[
d_k = (c_k, P_k, \hat{\mu}^{tip}_k, \hat{\Sigma}^{tip}_k, y_k, q_k).
\]
Here $c_k$ is the device family, $P_k$ is the estimated path, $\hat{\mu}^{tip}_k$ is the predicted tip location, $\hat{\Sigma}^{tip}_k$ is the tip covariance, $y_k$ is the placement label, and $q_k$ is an instance confidence.

\subsection{Fragment Detection and Association}

Rather than predict only at image level, the model first detects short device fragments and then groups them into instances. This compositional step lets it represent a variable number of devices, including overlapping or repeated devices from the same family. That matters in ICU radiographs, where several lines and tubes may be visible at once.

Let fragment $f_i$ have a local center, orientation, device-family logits, embedding vector, and tip offset distribution. Association constructs a fragment graph:
\[
G_F = (V_F, E_F),
\]
where nodes are fragments and edges represent geometric compatibility, embedding similarity, and path continuity. Connected components or graph clustering recover device candidates.

\subsection{Precision-Weighted Gaussian Tip Fusion}

Each fragment can predict a candidate tip location with uncertainty. For fragment $i$, let
\[
t_i \sim \mathcal{N}(\mu_i, \Sigma_i).
\]
For a device instance $d_k$ containing fragment set $F_k$, the device-level tip is fused by precision weighting:
\[
\hat{\Sigma}^{tip}_k =
\left(\sum_{i\in F_k} \Sigma_i^{-1}\right)^{-1},
\quad
\hat{\mu}^{tip}_k =
\hat{\Sigma}^{tip}_k
\left(\sum_{i\in F_k}\Sigma_i^{-1}\mu_i\right).
\]
Robust variants can use iteratively reweighted least squares with a Huber kernel to limit the effect of outlying fragments. The fused covariance is retained as graph evidence rather than discarded as an internal diagnostic.

\subsection{Placement Classification}

Placement classification is performed per device rather than only per image. For each instance $d_k$, the classifier estimates
\[
p(y_k \mid x, P_k, \hat{\mu}^{tip}_k, \hat{\Sigma}^{tip}_k, c_k).
\]
The graph stores both the categorical placement label and the score distribution. If a dataset supplies only image-level labels, a noisy-OR bridge can connect per-device predictions to image-level supervision, as in UCompCXR.

\subsection{Anatomical Reference Representation}

Device placement only makes sense relative to anatomy and image coverage. The graph therefore needs an anatomical reference layer: image borders, the carina when visible, diaphragms, lung fields, mediastinum, and an ``incomplete field-of-view'' state. When a landmark is absent or poorly seen, the model need not make a hard anatomical claim. It can instead store a landmark distribution $\mathcal{L}_j$ and visibility score $v_j$.

For a device-specific rule $R_c$ such as an expected relationship between a tube tip and a landmark, the system computes a margin distribution rather than a deterministic distance:
\[
m_{c,j} = h_c(\hat{\mu}^{tip}, \mathcal{L}_j), \qquad p(m_{c,j} \in R_c).
\]
The placement classifier can use this quantity, while the graph preserves it for review. If a rule cannot be evaluated because the relevant anatomy is outside the field of view, the appropriate output is ``indeterminate'' rather than normal or abnormal. This distinction is important when the image and report disagree.

\section{Empirical Foundation for the Visual Component}

The visual module uses UCompCXR, a compositional device-localization system \citep{lodhiya2026ucompcxr}. We added a deterministic graph-materialization pass over its saved predictions. Each study graph contains a \texttt{Study} node, \texttt{PredictedDevice} nodes, and edges to \texttt{TipEstimate}, \texttt{PlacementAssessment}, and, for B7, \texttt{FragmentProposal} nodes. The builder reads JSONL predictions and localization records only; it does not retrain a model, change a threshold, or alter a prediction.

The analysis covers the complete RANZCR CLiP test archive: 30,083 chest X-rays from 3,255 patients in five outer folds. We checked that each study and patient appears once across the test files. We report descriptive totals and patient-bootstrap percentile 95\% intervals from 2,000 resamples for device-field completeness. Because the test-plan lock was not enabled before the saved predictions were generated, this is a \emph{post-hoc descriptive} analysis rather than a confirmatory endpoint analysis. It demonstrates large-scale visual graph construction, not the text or longitudinal layers.

\begin{table}[H]
\centering
\caption{Visual prediction and graph-materialization results on the saved RANZCR CLiP test archive. B2 is the direct multi-task baseline and B7 is UCompCXR. Counts are exact across five non-overlapping outer folds; this post-hoc analysis is descriptive rather than confirmatory.}
\label{tab:visual-results}
\scriptsize
\begin{tabular}{@{}p{0.26\linewidth}p{0.33\linewidth}p{0.33\linewidth}@{}}
\toprule
Metric & B2 baseline & B7 UCompCXR \\
\midrule
Matched instances / false negatives / false positives & 11,660 / 5,649 / 93,992 & 14,721 / 2,588 / 23,854 \\
Detection sensitivity & 0.674 & 0.851 \\
Median tip error, \% image diagonal & 0.883 & 1.725 \\
Materialized graph nodes / edges & 718,351 / 688,268 & 914,632 / 884,549 \\
Predicted-device nodes & 344,134 & 118,647 \\
Median nodes per study [P10, P90] & 23 [9, 41] & 25 [19, 52] \\
Device nodes with covariance, placement probabilities, provenance, and fragment count & 0\% for each field & 100\% for each field \\
\bottomrule
\end{tabular}
\end{table}

\begin{figure}[t]
\centering
\includegraphics[width=\linewidth]{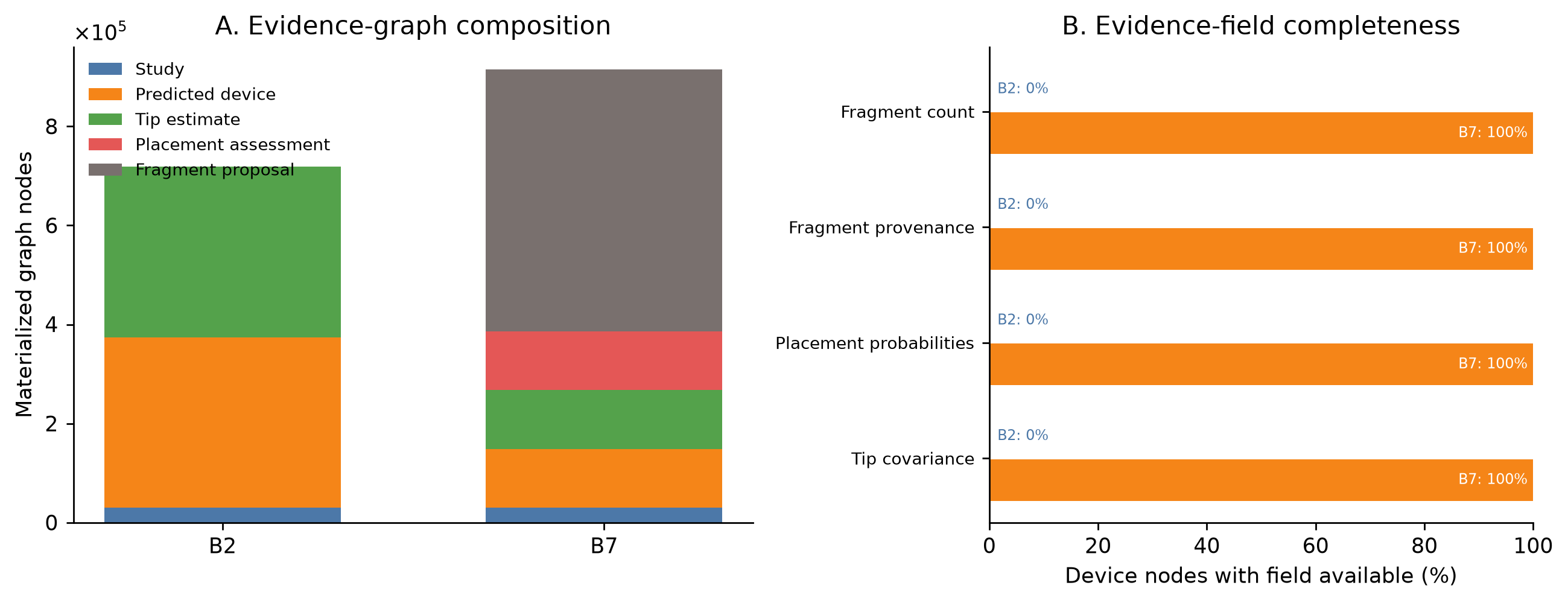}
\caption{Materialized evidence graphs from saved RANZCR CLiP test predictions. (A) B7 adds fragment-proposal and placement-assessment evidence while maintaining a comparable study-level graph size. (B) All B7 device nodes contain the four fields required for auditable visual evidence; B2 contains none. This is a post-hoc descriptive analysis of existing test predictions.}
\label{fig:materialized-graph}
\end{figure}

Figure~\ref{fig:materialized-graph} summarizes the graph composition. B7 produces fewer predicted-device nodes than B2, but it matches more devices (14,721 versus 11,660) and yields fewer false positives (23,854 versus 93,992). Every B7 device node carries the four fields needed to inspect a device-level assertion: a covariance-bearing tip estimate, placement-probability distribution, fragment provenance, and fragment count. B2 retains none of them. The patient-bootstrap intervals for B7 and B2 completeness were 100\% [100\%, 100\%] and 0\% [0\%, 0\%], respectively, for each field. These numbers describe field availability, not clinical calibration.

The B7 median tip error (1.725\% image diagonal) is higher than B2's 0.883\%, consistent with B7 recovering more difficult instances. It must therefore not be interpreted without its substantially higher detection sensitivity (0.851 versus 0.674). Nor do the materialization results demonstrate calibrated uncertainty after report grounding or cross-study linkage. Those properties require the multimodal evaluation in Section~\ref{sec:graph-eval}.

\section{Report and Event Extraction Module}

The text module turns reports and clinical notes into structured observations. It identifies device mentions, placement findings, recommendations, temporal cues, uncertainty language, and follow-up actions. For example:

\begin{itemize}
  \item ``ET tube tip projects 1.5 cm above the carina.''
  \item ``Enteric tube side port remains above the gastroesophageal junction.''
  \item ``Recommend advancing the nasogastric tube.''
  \item ``Right IJ central venous catheter tip terminates in the lower SVC.''
\end{itemize}

The extraction design follows the typed extraction and graph-storage pattern used in EventForge \citep{lodhiya2026eventforge}. A report event has the form:
\[
e = (type, device, assertion, anatomy, action, time, confidence, span).
\]
The system distinguishes factual observations, recommendations, negations, uncertainty expressions, and temporal references. Report extraction can be implemented with rule-assisted NLP, supervised models trained on radiology relation graphs, or typed LLM modules. For clinical deployment, extracted spans must be preserved so the graph remains auditable.

\section{Clinical Knowledge Graph Schema}

The graph contains image, report, device-instance, tip-estimate, placement-assessment, event, procedure, encounter, and patient nodes. Let
\[
G_C = (V_C, E_C, A_C),
\]
where $A_C$ stores attributes including uncertainty, provenance, timestamps, and access-control metadata.

\subsection{Node Types}

Core node types are:
\begin{itemize}
  \item \textbf{Patient}: de-identified patient identifier.
  \item \textbf{Encounter}: admission, ICU stay, or visit context.
  \item \textbf{Study}: chest X-ray study with timestamp and view.
  \item \textbf{Image}: individual image identifier and acquisition metadata.
  \item \textbf{Report}: radiology report text, sections, and version.
  \item \textbf{DeviceInstance}: detected catheter or tube instance.
  \item \textbf{TipEstimate}: mean, covariance, and calibration metadata.
  \item \textbf{PlacementAssessment}: normal, borderline, abnormal, incomplete, or unknown.
  \item \textbf{ReportEvent}: extracted finding, recommendation, or follow-up statement.
  \item \textbf{ProcedureEvent}: documented insertion, removal, advancement, retraction, or replacement.
\end{itemize}

\subsection{Edge Types}

Representative edge types are:
\begin{itemize}
  \item \texttt{HAS\_STUDY}: patient or encounter to imaging study.
  \item \texttt{HAS\_IMAGE}: study to image.
  \item \texttt{HAS\_REPORT}: study to report.
  \item \texttt{DETECTED\_IN}: device instance to image.
  \item \texttt{HAS\_TIP}: device instance to tip estimate.
  \item \texttt{ASSESSED\_AS}: device instance to placement assessment.
  \item \texttt{MENTIONED\_BY}: report event to report span.
  \item \texttt{GROUNDS}: report event to device instance.
  \item \texttt{RECOMMENDS}: report event to procedure action.
  \item \texttt{FOLLOWED\_BY}: event or study to later event or study.
  \item \texttt{POSSIBLY\_SAME\_DEVICE}: cross-study device linkage with confidence.
\end{itemize}

\section{Uncertainty Representation}

Uncertainty enters the graph in three forms: geometric, symbolic, and temporal.

\subsection{Geometric Uncertainty}

The visual module supplies geometric uncertainty directly:
\[
U_{geom}(d_k) = (\hat{\mu}^{tip}_k, \hat{\Sigma}^{tip}_k, coverage, calibration\_bin).
\]
The covariance ellipse can inform review triage. If a 95\% ellipse crosses an anatomical decision boundary, for example, the graph can flag the placement assessment as uncertain even when the categorical classifier is confident.

\subsection{Symbolic Uncertainty}

Symbolic uncertainty comes from extraction and linking:
\[
U_{sym}(e) = (p_{entity}, p_{relation}, p_{negation}, p_{temporal}, p_{grounding}).
\]
Phrases such as ``possibly'', ``may'', and ``suboptimally visualized'' reduce confidence. Negation and uncertainty detection matter because radiology reports are often deliberately qualified.

\subsection{Temporal Uncertainty}

Temporal uncertainty arises when device instances are linked across studies or a recommendation is aligned with a procedure note. A cross-study device link has probability:
\[
p(d_i \equiv d_j) =
\sigma(w_c C + w_t T + w_g G + w_r R),
\]
where $C$ is device-family compatibility, $T$ is temporal proximity, $G$ is geometric/path consistency, and $R$ is report-language consistency.

\subsection{Confidence Is Not Clinical Risk}

A model score is not the same as clinical risk. A high-confidence abnormal-placement score may still be unsuitable for automatic display when the field of view is incomplete, the device family is ambiguous, or the evidence is stale. We distinguish predictive confidence, geometric precision, evidence completeness, and workflow urgency. The first three are properties of the model or data; workflow urgency is a local, clinically governed policy decision.

Let $\rho(d)$ be a review score for device $d$. One conservative form is
\[
\rho(d) = \mathbb{I}[H(p(y\mid d)) > \tau_H]
+ \mathbb{I}[\operatorname{tr}(\hat{\Sigma}^{tip}) > \tau_\Sigma]
+ \mathbb{I}[\mathrm{conflict}(d)=1]
+ \mathbb{I}[\mathrm{coverage}(d)=0],
\]
where $H$ is predictive entropy and the indicator terms identify uncertain, conflicting, or incomplete-evidence cases. The score routes a case for review; it is not a diagnostic decision rule.

\section{Report-Image Alignment}

Report-image alignment links text events to visual device instances. Given report event $e$ and device instance $d_k$, the alignment score is:
\begin{align*}
A(e,d_k) ={}&
\alpha\,DeviceMatch(e,d_k)
+ \beta\,AnatomyMatch(e,d_k)\\
&+ \gamma\,PlacementMatch(e,d_k)
+ \delta\,TimeMatch(e,d_k)\\
&+ \eta\,UncertaintyCompat(e,d_k).
\end{align*}
The event is linked to the highest-scoring device only when the score clears a threshold; otherwise it remains ungrounded. This avoids forcing a link. A report may mention an incompletely imaged tube that the visual detector cannot confidently instantiate, and an image model may find a device the report does not mention.

\subsection{Constrained Grounding and Abstention}

Grounding is solved within a study unless the report explicitly refers to an earlier study. Candidate matches are first restricted by device family and document context. If a report mentions the same device family more than once, the matcher uses bipartite assignment so one visual instance cannot silently absorb every mention. Let $E_s$ be events and $D_s$ be visual instances in study $s$. The selected assignment $M$ maximizes
\[
\max_M \sum_{(e,d)\in M} A(e,d) - \lambda |M|
\]
subject to one-to-one matching within a compatible device family, with unmatched events and instances allowed. The penalty and acceptance threshold are calibrated on held-out annotated studies.

The output has four states: grounded, ungrounded-text, unmentioned-image, and ambiguous. ``Ungrounded-text'' means the report contains a device-related event but no adequate visual match exists. ``Unmentioned-image'' is a visual device without a corresponding report event. ``Ambiguous'' denotes more than one plausible match or inadequate evidence. These are first-class graph states, not errors to be hidden during data preparation.

\subsection{Conflict Detection}

A report-image conflict is not simply any difference between a report and a model. It becomes a review candidate only when both sources provide adequate, time-aligned, and comparable evidence. For example, a confident report finding of an enteric-tube side port above the gastroesophageal junction may conflict with an image-derived ``normal'' assessment. The graph retains both assertions and records a \texttt{POTENTIAL\_CONFLICT} edge with a reason code. It does not choose a winner automatically.

\section{Longitudinal Reasoning}

Temporal traversal supports longitudinal queries. For a device family $c$ in patient $p$, a trajectory is:
\[
T_{p,c} = \{(d_t, y_t, \hat{\mu}^{tip}_t, \hat{\Sigma}^{tip}_t, r_t)\}_{t=1}^{m},
\]
where $r_t$ includes report evidence. A malposition persistence query asks whether an abnormal or borderline assessment persists across studies:
\[
\exists t < t' : y_t \in \{abnormal,borderline\},\; y_{t'} \in \{abnormal,borderline\},\; p(d_t \equiv d_{t'}) > \theta.
\]

Recommendation follow-up can be expressed as a path query:
\begin{align*}
ReportEvent_{recommend} &\rightarrow ProcedureEvent_{action}\\
&\rightarrow Study_{followup} \rightarrow PlacementAssessment_{improved}.
\end{align*}
This representation supports questions beyond single-image accuracy: the proportion of abnormal tubes with documented follow-up, time to a corrected placement, or the number of high-uncertainty cases reviewed by a radiologist.

\subsection{Identity Management Across Studies}

Cross-study identity is the main source of longitudinal error. Temporally adjacent devices from the same family should not automatically be treated as the same device. Candidate links are restricted to an encounter-specific time window and scored using family, visible path topology, tip displacement, report statements about insertion or removal, and procedure events. A documented replacement or a long interval without visual continuity can break a chain.

The graph retains leading identity candidates and their probabilities instead of committing early to a single chain. Queries that claim persistent-device findings need a minimum linkage threshold. Case-finding queries may accept lower-confidence paths, but must show that uncertainty. This makes the distinction between exploratory analysis and safety-sensitive review explicit.

\subsection{Query Templates}

The following templates define the initial reasoning surface:
\begin{enumerate}
  \item \textbf{Persistence:} find device trajectories with consecutive abnormal or borderline assessments and a linkage probability above $\theta$.
  \item \textbf{Follow-up:} find recommendation events lacking a documented action or follow-up study within a configurable window; this is a record-completeness query, not evidence of missed care.
  \item \textbf{Resolution:} identify a recommendation, compatible corrective action, and subsequent assessment consistent with improvement.
  \item \textbf{Disagreement:} retrieve high-evidence report-image conflicts, grouped by device family and model version.
  \item \textbf{Uncertainty cohort:} retrieve cases whose geometric uncertainty, predictive uncertainty, or evidence completeness triggered a review flag.
\end{enumerate}

Every query returns its evidence paths, thresholds, and graph snapshot identifier with the result set. This gives a reviewer enough context to check the result and keeps downstream retrieval from treating graph outputs as unqualified truth.

\section{System Implementation}

A practical implementation can use PostgreSQL with pgvector for structured storage, dense retrieval, and similarity search. Visual outputs become structured rows; report spans and events retain their source offsets; and graph relations live in edge tables. For interfaces that query graph-backed clinical tables, schema-aware context pruning and execution-validated translation can reduce unnecessary schema exposure \citep{lodhiya2026schemaaware}. NetworkX-style processing is sufficient for offline graph analysis, while database indexes support query-time retrieval \citep{hagberg2008networkx}.

\begin{lstlisting}[style=pseudo]
for study in chest_xray_studies:
    image_outputs = visual_device_model(study.image)
    report_events = report_event_extractor(study.report)
    insert_study_nodes(study)
    insert_device_nodes(image_outputs)
    insert_report_event_nodes(report_events)
    link_report_events_to_devices(report_events, image_outputs)
    link_devices_across_prior_studies(study.patient_id, image_outputs)
    update_uncertainty_and_review_flags(study)
\end{lstlisting}

Minimum tables include:
\begin{itemize}
  \item \texttt{patients}, \texttt{encounters}, \texttt{studies}, \texttt{images}, \texttt{reports}
  \item \texttt{device\_instances(device\_id, study\_id, family, confidence)}
  \item \texttt{tip\_estimates(device\_id, x, y, cov\_xx, cov\_xy, cov\_yy, coverage)}
  \item \texttt{placement\_assessments(device\_id, label, probability, source)}
  \item \texttt{report\_events(event\_id, report\_id, type, span, confidence)}
  \item \texttt{graph\_edges(src, dst, relation, confidence, provenance)}
\end{itemize}

\subsection{Reference Architecture}

The reference deployment uses six bounded services: ingestion for de-identified inputs; visual inference; text extraction; a graph builder with typed contracts and versioned linkage rules; graph storage and a vector index; and a read-only review interface. Services communicate through immutable observation records. The graph builder may add edges, but it does not overwrite the raw report, original image identifier, or earlier model output. Corrections are written as new versioned observations.

Both batch research processing and event-driven updates fit this model. In batch mode, studies from an encounter can be processed in timestamp order. In event-driven mode, a finalized report or procedure note triggers an incremental update to existing candidates. A graph snapshot identifier ties each query result to the exact extraction models, calibration artifacts, ontology, and linkage policy that produced it.

\begin{table}[t]
\centering
\caption{Reference service boundaries and failure handling.}
\label{tab:services}
\small
\begin{tabular}{p{0.23\linewidth}p{0.33\linewidth}p{0.29\linewidth}}
\toprule
Service & Responsibility & Failure behavior \\
\midrule
Ingestion & Validate identifiers, time zones, de-identification, and source versions & Quarantine invalid records; do not create partial patient links \\
Visual inference & Emit device observations and calibration attributes & Emit ``model unavailable'' or ``indeterminate''; preserve no fabricated output \\
Text extraction & Emit span-grounded events and assertion attributes & Retain raw report; mark extraction unavailable or low confidence \\
Graph builder & Validate contracts, produce edges, retain provenance & Reject invalid edge; retain upstream observations for repair \\
Reasoning API & Run versioned graph queries and return evidence paths & Return incomplete-evidence state rather than an inferred clinical answer \\
Review interface & Present evidence and accept adjudications & Store adjudication as a new, attributed observation \\
\bottomrule
\end{tabular}
\end{table}

\subsection{Versioning and Human Adjudication}

Model outputs, rules, and annotations evolve. Each node and edge therefore carries \texttt{created\_at}, \texttt{valid\_from}, \texttt{valid\_to}, \texttt{producer\_id}, and \texttt{graph\_snapshot}. A human adjudication never deletes the automated output; it creates an attributed superseding assessment and records the reason for disagreement. This makes model-error analysis possible and avoids training on silently altered labels.

For research access, the implementation should maintain role-specific views. An analyst may query de-identified aggregate trajectories, while a clinical reviewer with appropriate authorization may inspect the source image and report. Free-text retrieval must respect the same access controls as the primary clinical record.

\section{Evaluation Protocol and Required Graph Evidence}
\label{sec:graph-eval}

The full system needs to be evaluated as a multimodal graph, not just as an image classifier. Table~\ref{tab:evaluation} lists the required evaluations. The visual results in Table~\ref{tab:visual-results} do not establish report grounding or longitudinal reasoning performance.

\begin{table}[t]
\centering
\caption{Evaluation matrix for uncertainty-aware clinical knowledge graphs.}
\label{tab:evaluation}
\begin{tabular}{p{0.23\linewidth}p{0.34\linewidth}p{0.32\linewidth}}
\toprule
Component & Task & Metrics \\
\midrule
Visual detection & Device instance detection and family classification & device F1, false positives per image, family accuracy \\
Tip localization & Tip mean and covariance quality & mean error, catastrophic error rate, 95\% coverage, calibration error \\
Placement assessment & Per-device normal/borderline/abnormal status & per-device AUROC, F1, sensitivity at fixed specificity \\
Report extraction & Findings, recommendations, uncertainty, temporal cues & entity F1, relation F1, span support, negation accuracy \\
Grounding & Link report events to device instances & grounding accuracy, top-k accuracy, abstention precision \\
Longitudinal graph & Link devices and events across studies & temporal edge F1, trajectory accuracy, correction-path accuracy \\
Clinical safety & Identify uncertain or conflicting cases & review triage precision, conflict detection, calibration under shift \\
\bottomrule
\end{tabular}
\end{table}

\subsection{Datasets}

The image side can begin with RANZCR CLiP because it contains catheter/tube labels and traced device paths for a subset of images \citep{ranzcr2021clip}. Report and longitudinal evaluation can use report-bearing chest X-ray datasets such as MIMIC-CXR where appropriate permissions and de-identification rules apply \citep{johnson2019mimiccxr}. A full graph benchmark would require derived labels for report-device grounding, recommendation follow-up, cross-study device linkage, and corrective-action events.

\subsection{Benchmark Construction}

All studies from one patient should stay in one partition. Image-level random splits would leak longitudinal context and inflate temporal-linkage estimates. The development set selects calibration and abstention thresholds; the test set stays untouched until matching and query logic are fixed. Where data-use agreements permit it, external validation should also vary institution, scanner population, and report style.

Annotation should be staged rather than asking annotators to label an entire graph at once. First, annotators identify device mentions, assertion categories, and supporting spans. Second, they link report events to device instances or explicitly select ``no visual match'' and ``ambiguous''. Third, they annotate cross-study identity only over temporally adjacent candidate pairs, with access to documented insertion or removal events when available. Finally, adjudicators label a focused set of longitudinal questions such as persistence, documented follow-up, and report-image disagreement. This staged approach produces reusable component labels and makes disagreement measurable.

The benchmark must distinguish absent information from negative findings. No report mention of a device is not evidence that the device is absent, and no documented follow-up action is not evidence that no action occurred. These distinctions should be encoded as annotation states and reported in label-prevalence tables.

\subsection{Metric Definitions}

For tip uncertainty, empirical coverage at nominal level $\alpha$ is
\[
\mathrm{Cov}_{\alpha}=\frac{1}{N}\sum_{i=1}^{N}
\mathbb{I}\left[(t_i-\hat{\mu}_i)^\top\hat{\Sigma}_i^{-1}(t_i-\hat{\mu}_i)
\leq \chi^2_{2,\alpha}\right].
\]
Calibration should report coverage at multiple levels, not only a single 95\% number. For categorical placement assessments, report expected calibration error together with reliability plots and class-specific sensitivity. For grounding and temporal linkage, report both precision at accepted links and selective risk as a function of coverage. An abstaining system should be compared at matched coverage; otherwise an apparent accuracy gain may be achieved simply by declining difficult cases.

Longitudinal queries should be evaluated as structured predictions. A persistence query is correct only if the component device links and all qualifying assessments are correct. A follow-up query is correct only if the temporal window, action type, and evidence provenance match the adjudicated path. Report task-level precision, recall, F1, and exact evidence-path accuracy separately; a query answer without the correct supporting path is insufficient for auditability.

\subsection{Statistical Reporting}

All primary metrics should include patient-bootstrap confidence intervals. Comparisons between system variants should use paired resampling at the patient level, because studies from the same encounter are correlated. The evaluation report should predefine primary endpoints, error-taxonomy categories, exclusions, and handling of missing reports or unavailable prior studies. Where a threshold is tuned, the manuscript should identify the development cohort and state that the threshold was frozen before test evaluation.

\subsection{Baselines}

Baselines should include image-level placement classification, segmentation plus global placement classification, report-only extraction, text-only RAG over reports, graph extraction without image uncertainty, and the full uncertainty-aware graph. Ablations should remove covariance-aware tip representation, report grounding, temporal device linkage, and uncertainty-based review flags.

\subsection{Hypotheses}

The main hypotheses are:
\begin{enumerate}
  \item Instance-level graph representation improves device-specific reasoning relative to image-level labels.
  \item Covariance-aware tip nodes improve review triage and reduce unsupported placement certainty.
  \item Report-image grounding improves longitudinal follow-up queries relative to report-only extraction.
  \item Graph trajectories expose clinically important persistence and correction patterns not visible in single-study evaluation.
\end{enumerate}

\subsection{Ablation and Stress-Test Plan}

The minimum ablation suite compares the full system with: (i) image-only per-device inference, (ii) text-only report extraction, (iii) a graph without tip covariance, (iv) a graph without abstention, and (v) a graph without cross-study identity links. Stress tests should stratify by device family, portable versus nonportable acquisition, image quality, multiple-device cases, missing reports, changed report templates, and out-of-distribution institutions. A safety-oriented error analysis should separately count false reassurance, unsupported longitudinal linking, ungrounded recommendation matching, and missed uncertainty-routing opportunities.

\subsection{Scope of Results}

The manuscript includes a reproducible, large-scale visual evidence-graph materialization study over a complete five-fold held-out prediction archive. It is not an end-to-end multimodal benchmark: radiology reports, recommendations, procedures, and longitudinal links were not evaluated. The test analysis is post-hoc descriptive because the original plan lock was disabled before prediction generation, so it is not a preregistered confirmatory endpoint. The next empirical revision should freeze the full pipeline before a new test evaluation and add cohort flow, annotation agreement, report-grounding and temporal-linkage results, confidence intervals, and representative error analyses.

\section{Reproducibility Checklist}

A future implementation release should include a versioned graph schema and ontology, JSON schemas for each observation type, deterministic graph-builder code, model cards for visual and text components, calibration artifacts, threshold-selection notebooks, patient-level split manifests where redistribution is allowed, annotation guidance, query templates with expected evidence paths, and a governance statement for each dataset. Images and reports that cannot be redistributed should be referenced through dataset-approved identifiers and scripts that run in an authorized environment.

Each reported experiment should log a configuration hash that includes the visual model checkpoint, text extractor, ontology version, linkage weights, calibration version, and graph builder commit. This is particularly important for multimodal systems, where a small change in a linker can affect downstream quality-improvement counts without changing image-model performance. The graph snapshot identifier should appear in every result table and error-analysis export.

\section{Clinical Safety and Governance}

This system is decision support, not autonomous diagnosis. It should preserve source spans, image coordinates, uncertainty values, model versions, and timestamps. High-uncertainty predictions should be visible, not hidden. The audit trail should show why a recommendation was linked to a device, which report span supports a finding, and whether a later study contradicts an earlier assessment.

Access control and de-identification are mandatory for any deployment using patient data. Research datasets must be handled under their licenses and data-use agreements. Longitudinal linkage should use de-identified patient and encounter keys. Any clinical use requires prospective validation, workflow testing, and human oversight.

The system must also be evaluated for distributional and documentation bias. Device visibility, report detail, and procedure documentation can differ across care settings and patient populations. A graph may amplify those differences if missing documentation is mistaken for a negative event or if a calibration threshold is transferred without validation. Monitoring should therefore include missingness rates, accepted-link rates, abstention rates, calibration, and error categories by site, device family, acquisition context, and other clinically justified subgroups. These analyses support safety monitoring; they do not substitute for an equity assessment designed with domain experts.

\section{Limitations}

The visual evidence-graph study is real but narrow. It materializes model outputs from RANZCR CLiP, which lacks report text, procedures, recommendations, and a longitudinal timeline for testing the broader graph. The post-hoc test analysis is not confirmatory because the original lock was disabled before the saved predictions were made. A retained covariance or provenance field shows that evidence is available; it does not show that uncertainty is calibrated, that a placement assessment is clinically correct, or that an automated conclusion is safe. The complete system needs report-device grounding labels, longitudinal follow-up annotations, an independently frozen evaluation plan, and clinical review. Ambiguous language can derail report extraction; cross-study linking can confuse replacement with persistence; records may omit text events; and uncertainty calibration can shift across scanners, institutions, and patient populations. The graph is intended to support review and research, not infer unrecorded care.

\section{Conclusion}

Chest X-ray device assessment needs more than per-image labels. We show that a complete RANZCR CLiP test-prediction archive can be converted into a large, typed visual evidence graph while retaining device instances, tip uncertainty, placement distributions, and fragment provenance. The same evidence model can incorporate report language, recommendations, procedures, and longitudinal follow-up without losing the sources behind each assertion. A frozen, report-bearing patient-level benchmark with grounded multimodal and temporal endpoints is needed before drawing conclusions about those extensions. The present results support an auditable visual evidence graph; they do not establish a validated clinical decision system.

\bibliographystyle{unsrtnat}
\bibliography{references}

\end{document}